\documentclass[conference, table]{IEEEtran}
\IEEEoverridecommandlockouts

\usepackage{cite} 
\usepackage{amsmath,amssymb,amsfonts}
\usepackage{algorithmic}
\usepackage{graphicx}
\usepackage{textcomp}
\usepackage{epsfig} 
\usepackage{times} 
\usepackage{nicefrac}
\usepackage{bm}
\usepackage{enumerate}
\usepackage{caption}
\usepackage{subcaption} 
\usepackage{float}
\usepackage[table,xcdraw]{xcolor}
\usepackage{tabularx}
\usepackage{cancel}
\usepackage{caption}

\def\BibTeX{{\rm B\kern-.05em{\sc i\kern-.025em b}\kern-.08em
    T\kern-.1667em\lower.7ex\hbox{E}\kern-.125emX}}

\graphicspath{ {./images/} }
\begin{document}


\title{\LARGE \bf
    Risk Averse Bayesian Reward Learning for Autonomous Navigation from Human Demonstration
    \vspace{-2ex}
}

\author{
    Christian Ellis$^{1}$, Maggie Wigness$^{2}$, John Rogers$^{2}$, Craig Lennon$^{2}$, and Lance Fiondella$^{1}$ \\
    
    \setcounter{figure}{0} 
    \begin{minipage}[c]{\textwidth}
        \centering
        \includegraphics[width=\textwidth]{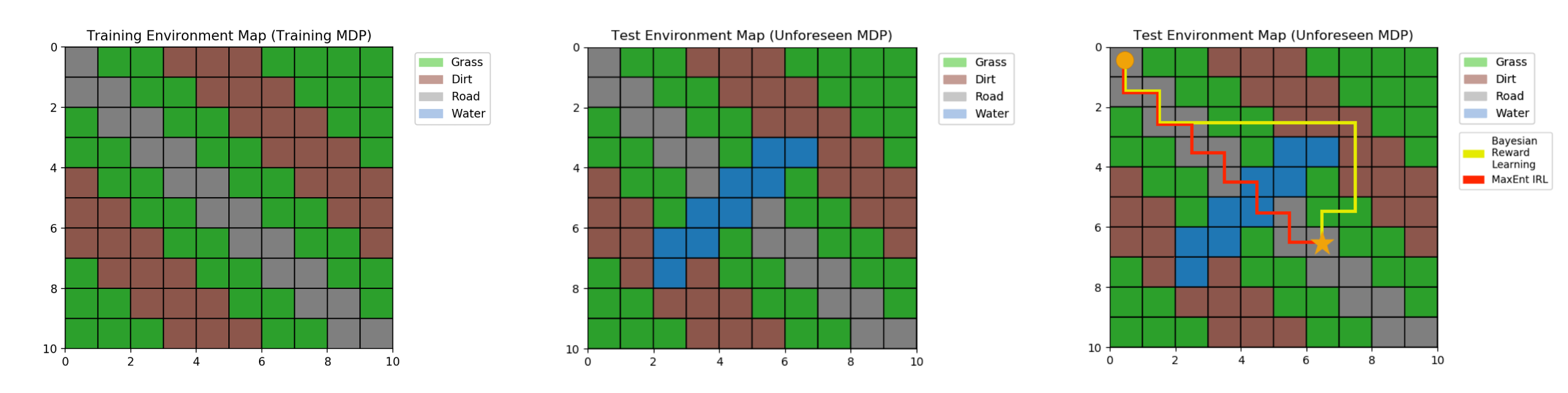}
        \captionof{figure}{Simplified illustration to demonstrate the impact of distributional shift. Left: Training environment. Middle: Test environment. Right: Planned trajectories in the testing environment for various learning methods.}
        \label{fig:gridworld}
    \end{minipage}

    \thanks{*This work was supported by the Army Research Laboratory.}%
    \thanks{
        $^{1}$
        Christian Ellis is a PhD Student and Lance Fiondella is an Associate Professor in the Department of Electrical and Computer Engineering at the University of Massachusetts Dartmouth, USA.
    }
    \thanks{
        $^{2}$
        Maggie Wigness, John Rogers, and Craig Lennon are researchers at the United States Army Research Laboratory (ARL).
    }
}
\maketitle

\begin{abstract}
Traditional imitation learning provides a set of methods and algorithms to learn a reward function or policy from expert demonstrations.
Learning from demonstration has been shown to be advantageous for navigation tasks as it allows for machine learning non-experts to quickly provide information needed to learn complex traversal behaviors.
However, a minimal set of demonstrations is unlikely to capture all relevant information needed to achieve the desired behavior in every possible future operational environment.
Due to distributional shift among environments, a robot may encounter features that were rarely or never observed during training for which the appropriate reward value is uncertain, leading to undesired outcomes.
This paper proposes a Bayesian technique which quantifies uncertainty over the weights of a linear reward function given a dataset of minimal human demonstrations to operate safely in dynamic environments.
This uncertainty is quantified and incorporated into a risk averse set of weights used to generate cost maps for planning.
Experiments in a 3-D environment with a simulated robot show that our proposed algorithm enables a robot to avoid dangerous terrain completely in two out of three test scenarios and accumulates a lower amount of risk than related approaches in all scenarios without requiring any additional demonstrations.

\end{abstract}

\section{INTRODUCTION}
Robot behavior can be described through a reward function which directs the robot toward states leading to the achievement of goals and developer specifications~\cite{kober_reinforcement_2013}.
This encoding often focuses on goals defined during system design, implicitly expressing indifference to all others, resulting in negative side effects~\cite{amodei_concrete_2016} such as the robot traversing a harmful terrain or crashing into an obstacle.
Inverse reinforcement learning  (IRL)~\cite{ng_algorithms_2000,abbeel_apprenticeship_2004,ziebart2008maximum} seeks to learn a reward function given a dataset of demonstrations, which eliminates the need for an expert to hand code these rewards.
However, the performance of these approaches are sensitive to the features seen in the demonstrations.
Certain states and the features that describe them may have never been encountered, leading to uncertainty in their reward~\cite{leike_ai_2017, hadfield-menell_inverse_2017}.
This presents a challenge for autonomous navigation since it is desirable for a robotic system to be robust to operation in unknown, possibly adversarial, environments which are likely to contain unforeseen conditions.

As an example, consider an autonomous ground robot which learns to navigate in an environment consisting of four types of terrain, including grass, dirt, road, and water (Fig. \ref{fig:gridworld} middle).
In the training environment (Fig. \ref{fig:gridworld} left), none of the states contain the water feature, and subsequently neither do the demonstrations.
Maximum likelihood approaches such as~\cite{ziebart2008maximum} implicitly set the reward weight for water to their initialization value because its feature count is zero.
Since the demonstrations do not provide full reward information, the robot fails to avoid water in the test environment (Fig. \ref{fig:gridworld} right).
Water may have been avoided if different initial rewards were used, but this would not resolve the more general problem of how to handle states possessing uncertain reward values.
As a step toward solving this problem, we propose a systematic approach which utilizes Bayesian analysis to quantify uncertainty for each terrain's reward weight.
This approach allows the robot to achieve risk averse behavior by avoiding terrain possessing high uncertainty (Fig. \ref{fig:gridworld} right).

Previous imitation learning methods have provided a Bayesian framework to incorporate prior information and obtain a unique reward function ~\cite{ramachandran2007bayesian, choi2011map}.
Most similar to our work, Hadfield-Menell et al.~\cite{hadfield-menell_inverse_2017} provide a Bayesian technique, which explicitly considers safety to distributional shift in environments.
The difference between their methodology and ours is the assumption of a Bayesian posterior.
Specifically, they obtain a posterior over reward functions given a proxy reward function and a world model, while we do not assume a proxy reward function is given, but rather a dataset of demonstrations.
This change in assumptions allows developers to obtain a reward function which explicitly considers safety solely from observed behavior.
However, the main difference between Ref. \cite{hadfield-menell_inverse_2017} and our methodology is in the reward function selection technique.
The reward function obtained in Ref. \cite{hadfield-menell_inverse_2017} changes at runtime, requiring planning with multiple reward functions online, whereas our technique outputs a single reward function at the end of training.
Online planning with a single reward function reduces computational overhead, thereby increasing the capability to scale to larger environments.

We build upon previous IRL work \cite{wigness_robot_2018}, which learned traversal behavior reward models for autonomous navigation. This method assigns high rewards to state features visited more frequently during demonstration, implicitly assuming that high visitation frequency means the state feature is better than those visited less frequently. Our inclusion of uncertainty modifies this assumption such that if a state feature is visited less it is associated with more uncertainty.
We propose \emph{risk averse Bayesian reward learning (RABRL)}, as a method to obtain a unique, risk averse linear reward function solely from a dataset of demonstrations for autonomous waypoint navigation.
The contributions are twofold, we provide a methodology to quantify uncertainty over reward functions from a set of human demonstrations, and provide a weight selection technique, which chooses a single weight set during training.

The remainder of the paper is organized as follows. Section~\ref{sec:related_work} summarizes relevant related work. Section~\ref{sec:methodology} presents our methodology showing how a posterior over reward functions is used to obtain a unique set of risk averse weights. Section~\ref{sec:experiment} describes experimental setup and results obtained from an autonomous ground robot navigating in a 3-D simulation. Section~\ref{sec:conclusion} concludes with areas this research can impact and identifies opportunities for future extensions. 

\section {RELATED WORK}\label{sec:related_work}
RABRL seeks to obtain risk averse navigation behavior by learning a reward function from minimal human demonstrations. Toward this end, this paper incorporates imitation learning, Bayesian IRL, and safe autonomy.
A brief overview of each follows.

\subsection{Learning from Demonstration}
In reward learning, the goal is to obtain a unique reward function from human demonstration that maps trajectories to scalar rewards~\cite{jeon2020reward}.
More generally, autonomous agents learning solely from demonstrations to replicate behavior is called imitation learning~\cite{osa_algorithmic_2018}.
There are two distinct sub-fields within imitation learning, behavioral cloning~\cite{bain1995framework, garret_bc} and IRL~\cite{ng_algorithms_2000}. A robot learns a policy directly in behavioral cloning, whereas a robot implementing IRL learns a reward function, which may then be used to obtain a policy.

IRL is an ill-posed problem~\cite{ng_algorithms_2000} because many possible reward functions can characterize robot behavior.
Attempts to identify effective solutions have led to several competing methodologies. For a comprehensive survey,  we refer the reader to the following references \cite{osa_algorithmic_2018, arora2018survey, Zhifei012survey, argall2009survey}.
Nevertheless, previous work shows that learning from demonstration scales to real robotic systems for both linear~\cite{ratliff2006maximum, wigness_robot_2018}, and non-linear problems~\cite{wulfmeier2016watch}.
Similarly, deep learning approaches to IRL have been successful in the Atari~\cite{ibarz2018reward} and MuJoCo environments~\cite{brown2019extrapolating}.
Since we seek to learn a representative reward function from minimal demonstrations, we employ a linear model instead of a nonlinear deep-learning approach.

\subsection{Bayesian IRL}
Bayesian IRL methods provide a subjective approach to reason about many different reward functions, assigning each a point probability.
Ramachandran and Amir~\cite{ramachandran2007bayesian} proposed Bayesian IRL as a methodology to build a posterior density over reward functions given a dataset of demonstrations.
Choi and Kim~\cite{choi2011map} developed a framework subsuming previous IRL methods and showed that the maximum a posteriori estimator is a better estimator than the posterior mean proposed in \cite{ramachandran2007bayesian}.
Our paper also takes the Bayesian viewpoint to enable the robot to quantifiably establish a belief over multiple reward functions and evaluate their uncertainty.
In contrast to Refs. \cite{ramachandran2007bayesian, choi2011map}, our methodology differs because we explicitly consider safety when learning from demonstration.

\subsection{Safe Autonomy}
The reward function obtained from demonstrations in a training environment may not be well suited for guiding robot behavior in a new operational environment, leading to negative side effects\cite{amodei_concrete_2016, leike_ai_2017}.
Safe imitation learning seeks to obtain behavior that avoids negative side effects \cite{zhang_safedagger, menda_ensembledagger, brown2020safe}.
Although these approaches explicitly consider safety, they do not directly address changes in the environment (distributional shift).
Lutjens et al. \cite{lotjens2019safe} obtained model uncertainty estimates to avoid novel obstacles from perception systems in the reinforcement learning framework, but did not consider the imitation learning scenario.
Janson et al. \cite{janson2018safe} sought safe motion planning in unknown environments, but only considered obstacle avoidance and not a preference over different terrains.
Hadfield-Menell et al.~\cite{hadfield-menell_inverse_2017} considered safety to distributional shift, but do not explicitly consider learning from demonstration and their weight selection technique changes at runtime.
Our intended use case is one in which training occurs from a dataset of demonstrations in an environment that is likely to be different than the operational environment. Consequently, we seek a method to obtain a single reward model
before operation in a new environment, both for traceability and to save time and computational resources, when operating online in the deployment environment.

\section{METHODOLOGY}\label{sec:methodology}
An outline of the methodology follows.
Section \ref{sec:meth:agent_model} formally models the problem as an MDP without rewards. 
Section \ref{sec:meth:prob_form} formulates the problem.
Section \ref{sec:meth:likelihood} explains how the likelihood of a demonstrator's reward intent over terrains is modeled.
Section \ref{sec:meth:prior} describes two distributions to encode prior reward information.
Finally, section \ref{sec:meth:weight_selection} describes our methodology to select the reward weights for each feature.

\subsection{Environment and Robot Modeling} \label{sec:meth:agent_model}
An autonomous robot navigating in an environment is modeled using a Markov Decision Process (MDP), $M$, represented by the following tuple
\begin{equation}
\label{eq_MDP}
    M = \langle \mathcal{S}, \mathcal{A}, T, R, \gamma \rangle
\end{equation}
where $\mathcal{S}$ is a set of states, $\mathcal{A}$ is a set of actions, $T$ is the state transition distribution over the next state given the present state and action represented by $T(s_{t+1}|s_t, a)$, $R$ is the reward function representing the numerical reward received by taking action $a \in A$ in state $s \in S$ represented by $R(s, a): \mathcal{S} \times \mathcal{A} \rightarrow R$, and $\gamma \in [0,1)$ is the discount factor representing the weight on future unseen rewards. The solution to an MDP is a policy $\pi(s) : \mathcal{S} \rightarrow \mathcal{A}$, determining the action $a$ a robot will take in state $s$. The optimal policy for an MDP ($\pi^*$)  maximizes the expected cumulative reward.

Scenarios where the agent finds itself in an unknown environment are modeled by omitting $R$ in Eq.~(\ref{eq_MDP}).
\begin{equation}
\label{mdp_norewards}
    M \backslash R = \langle \mathcal{S}, \mathcal{A}, T, -, \gamma \rangle
\end{equation}
In the reinforcement learning framework, a reward function is considered to be the most succinct, robust, and transferable definition of a task \cite{ng_algorithms_2000}.
To learn a reward function, an agent is supplied demonstrations in the form of trajectories that depict the desired behavior.
For a navigation task, the cumulative reward associated with a trajectory demonstration can be found through its state sequence $\xi = [s_1, s_2, ..., s_T]$, where $T$ is the number of time steps.
Although we only consider reward functions that are a function of state $R(s_t)$, one may also wish to consider reward functions that are a function of the state and action $R(s_t, a_t)$ or reward functions that are a function of the state, action, and next state $R(s_t, a_t, s_t^\prime)$.
This modeling choice depends on the goals and preferences the system designer desires the agent to learn~\cite{sutton_reinforcement_2018}. 
Formally, this is an example of a reward design problem \cite{singh2009rewards}, where the true reward function is unobservable, but possible reward functions are assessed by a fitness function given a distribution of environments the agent may find itself in.
The behavior of an agent operating in $M \backslash R$ is summarized as a probability distribution over trajectories given a vector of reward weights $P(\xi | \hat{w})$ and is referred to as a robot model.
For a list of robot models incorporating human feedback, we refer the reader to \cite{jeon2020reward}.

\subsection{Problem Formulation} \label{sec:meth:prob_form}
We seek a posterior over the weights describing a reward function,
\begin{equation}
  \label{eq_posterior}
  \begin{split}
      P(w = \hat{w} | \mathcal{D}) = \frac{P(\mathcal{D} | \hat{w}) P(\hat{w})}{P(\mathcal{D})}
  \end{split}
\end{equation}

\noindent where $w$ is a random vector describing the weights of a reward function, $\hat{w}$ is an estimator of $w$,  $\mathcal{D}$ is a dataset of demonstrations (navigation trajectories) such that $\mathcal{D} = \{(\xi_i), (\xi_{i+1}), \dots, (\xi_{n})\}_{i=1}^{n}$.

We consider reward functions as a function of trajectory states expressed as a linear combination between estimated weights $\hat{w}$ and features $\phi(s)$ representing a state such that $\phi: \mathcal{S} \rightarrow \mathbb{R}^\mathbb{D}$, where $\mathbb{D}$ indicates the dimension of the feature space.
\begin{equation}\label{state_reward)}
   R(s) = \hat{w}^T\phi(s)
\end{equation}

\noindent The total reward for a trajectory is the sum of its state rewards.
\begin{equation}\label{total_return}
   R(\xi) = \sum_{s_i \in \xi} \hat{w}^T\phi(s_i)
\end{equation}

We define the reward space $\mathcal{R}$ is a discrete set of fixed weight vectors, which parameterize a reward function.
The discrete set $\mathcal{W}$ contains all possible values for a single element in the weight vector.
The number of possible weight vectors is therefore represented by the cardinality of $\mathcal{W}$ raised to the dimension of the feature space, $|\mathcal{R}| = |\mathcal{W}|^\mathbb{D}$.
The reward space $\mathcal{R}$ should contain values representative of the number of features and their scaled differences. At a minimum, $|\mathcal{W}|$ should be equal to $\mathbb{D}$, so that each feature may be assigned a distinct value, representing the preference over features. However, to enable a reward model to consider the scaled reward difference, such as ``water is 10 times worse than grass,'' a larger or non uniformly spaced set of values can be specified to capture such preference.
The training time of the model increases as $|R|$ and $|\mathcal{S}|$ increase, so it is important to choose a value of $|R|$ relative to the complexity of the domain. The posterior is computed at $|\mathcal{R}|$ discrete points to obtain a nominal probability for each point. These nominal points are subsequently divided by their marginal probability producing a valid discrete posterior distribution.

Alternatively, with the use of Markov Chain Monte Carlo \cite{andrieu2003introduction}, a continuous posterior over reward functions can be obtained, allowing one to model an infinite number of reward functions.
However, this requires numerically approximating integrals, adding computational complexity during training. Furthermore, in the context of our weight selection technique (Section \ref{sec:meth:weight_selection}), a continuous posterior would require using differential entropy, which is difficult to interpret \cite{michalowicz2013handbook}.
Due to these limitations and the performance achieved with a small number of reward functions (Section \ref{sec:eval:results}), we chose the discretization approach.

\subsection{Likelihood Modeling} \label{sec:meth:likelihood}
The likelihood of a demonstrator assuming independent and identically distributed trajectories is defined as the product of the individual trajectories.
\begin{equation}
    \begin{aligned}
        P(\mathcal{D} | \hat{w}) & = P(\xi_1 | \hat{w}) \times P(\xi_2 | \hat{w}) \times ... \times  P(\xi_n | \hat{w}) \\ 
        & = \prod_{\xi_i \in \mathcal{D}} P(\xi_i | \hat{w})
    \end{aligned}
\end{equation}
More specifically, we model the demonstrator using the maximum entropy IRL distribution \cite{ziebart2008maximum}.

\begin{equation} \label{eq_posterior_likelihood}
    \begin{split}
        P(\mathcal{D} | \hat{w}) \propto \prod_{\xi_i \in \mathcal{D}} \exp{(\beta \hat{w}^T \mathbb{E}[\phi(\xi_i) | \xi_i \sim P(\xi_i | \hat{w})])}
    \end{split}
\end{equation}

\noindent where $\beta\in [0, 1]$ represents the level of confidence in a demonstrator. $\beta = 0$ indicates low confidence, such as random behavior from a demonstrator, while $\beta = 1$ indicates optimal behavior with respect to the reward preference over features.
The expected feature count is high when a trajectory $\xi_i$ obtains high rewards for a given robot model $P(\xi | \hat{w})$ relative to all other trajectories and vice versa, as represented by the dot product between weights and the expected feature count.
Therefore, an increase in the reward increases a weight vector's desirability, quantifying a preference over features with respect to $\mathcal{D}$.

The feature expectation of all demonstrated trajectories is:
\begin{equation} \label{eq:feat_expectations}
  \mathbb{E}[\phi(\mathcal{\xi})] = \sum_{\xi_i \in \bm{\xi}}  P(\xi_i | \hat{w}) \phi(\xi_i)
\end{equation}
Where $\bm{\xi}$ represents the set of all possible trajectories that can be taken in the MDP. 

Although several candidate robot models may be suitable \cite{jeon2020reward}, we use maximum entropy IRL~\cite{ziebart2008maximum}, which also contains an algorithm to compute Eq.~(\ref{eq:feat_expectations}).
\begin{equation} \label{eq_maxent_likelihood}
  P(\xi | \hat{w}) = \frac{\exp(\hat{w}^T \phi(\xi))}{Z(\xi)}
\end{equation}
Calculating the feature expectation directly is infeasible because it requires an agent to consider all possible trajectories in the MDP, as captured in the normalization constant $Z(\xi)$.
However, this can be approximated by using either the forward backward algorithm or value iteration~\cite{ziebart2010modeling}.

\subsection{Prior Modeling} \label{sec:meth:prior}

There are multiple ways to incorporate prior information about reward weights, $P(\hat{w})$. For our work, we chose a modified uniform prior as an uninformative prior, and a Dirichlet prior as an informative prior.

If a demonstrator does not prefer any one weight paramaterization, a uniform prior may be used.
However, if a reward function has all the same weights for each feature, any set of demonstrations appear Boltzmann optimal \cite{ng_algorithms_2000}.
Therefore, we use a modified version of the discrete uniform prior, where all weight sets have equal probability \emph{unless} all its weights are the same, in which case its probability is zero.
\begin{equation} \label{discrete_uniform_prior}
P(\hat{w}) = 
\begin{cases} 
      0 & \text{if, } \hat{w}_1 = \hat{w}_2 = ... = \hat{w}_n \\ 
      \frac{1}{|\mathcal{R}| - |\mathcal{W}|} & \text{otherwise}
   \end{cases}
\end{equation}
\noindent Each element in the weight vector $\hat{w}$ above takes a value from the discrete set $\mathcal{W}$.

In some cases, a preference over terrains is known a priori, and therefore can be captured by a Dirichlet prior, a continuous multivariate generalization of the beta distribution
\begin{equation} \label{dirichlet_prior}
    P(\hat{w}) = \frac{1}{Beta(\alpha)} \prod_{i=1}^{\mathbb{D}}\hat{w}_i^{\alpha_i-1} \quad \forall i, \,  1< \alpha_i   
\end{equation}
such that $\alpha \in \mathbb{R}_{+}^\mathbb{D}$ where each $\alpha_i$ represents our preference over a corresponding feature weight $w_i$.
The higher the $\alpha_i$, the more density the component possesses.
That is, a large value of $\alpha_i$, corresponds to preference over all other components $j$ for which $\alpha_i > \alpha_j$ is true.
The Dirichlet distribution is subject to the constraint $\sum_{i=1}^{|\mathcal{W}|} w_i = 1$.
To satisfy this constraint, a softmax function is applied to the current weight vector, producing normalized weights.
\begin{equation} \label{softmax}
    \hat{w}_i \leftarrow \frac{\exp(\hat{w}_i)}{\sum_{j=1}^{|\mathcal{W}|} \exp(\hat{w}_j)}
\end{equation}
In the Dirichlet prior, $\hat{w}$ is assumed to be continuous, while in Sec.~(\ref{sec:meth:agent_model}) we have defined it to be discrete. However, a proper prior is obtained as a result of applying Eq.~(\ref{softmax}) to a given weight vector before obtaining the result in Eq.~(\ref{dirichlet_prior}).

\subsection{Planning Risk Averse Behavior}
\label{sec:meth:weight_selection}
A Bayesian posterior is obtained by evaluating Eq.~(\ref{eq_posterior}) for $|\mathcal{R}|$ different reward vectors.
A point evaluation is obtained by multiplying the likelihood (Eq.~(\ref{eq_posterior_likelihood})) and a prior (Eq.~(\ref{dirichlet_prior})) or (Eq.~(\ref{discrete_uniform_prior})) for each reward function in $\mathcal{R}$. Then, a normalization constant is computed by taking the sum of the point products.
The system designer can then quantify uncertainty over $\hat{w}$.
Uncertainty is expressed as the normalized Shannon entropy of the marginal probability distribution of each feature weight $\hat{w_i} \in \hat{w}$.
The marginal probability is calculated by holding the weight being marginalized constant and summing over all possible values of the other $n-1$ variates.
\begin{equation} \label{eq:marginal}
        p_{w_i}(k) = \sum_{\forall \hat{w_j} \in \hat{w} | \hat{w_j} \neq \hat{w_i}, \hspace{0.3em} \hat{w_j} \in |\mathcal{W}|} P(w_1, \dots, w_i = k, \dots, w_n)
\end{equation}

Uncertainty is defined according to each individual feature, since each corresponds to a distinct semantic meaning. The overall uncertainty of each marginal distribution is quantified by the normalized Shannon entropy
\begin{equation}\label{shannon_entropy} 
    H(w_i) = -\frac{\sum_{k=1}^{|\mathcal{R}|}p_{w_i}(w_k) \log_2 p_{w_i}(w_k)} {\log_2|\mathcal{R}|}
\end{equation}
yielding a value in the interval $[0,1]$.
An entropy of 0 indicates certainty in the weight's value, and a value of 1 represents maximum uncertainty, the uniform distribution.
Weights are then chosen by their respective uncertainty using a specified level of acceptable risk, $\epsilon$, where smaller values of $\epsilon$ indicate greater risk acceptance.
When the entropy is greater than or equal to the threshold, $ H(w_i) \geq 1 - \epsilon$, the lowest reward weight is chosen.
Otherwise, if $ H(w_i) < 1 - \epsilon$, the expected value $E[w_i]$ of the marginal distribution is chosen as the reward weight.

The weights of the linear reward function are then used to produce a costmap for a navigation planning algorithm.
Costmap generation is a simple dot product between the reward weight vector and the environment feature maps (discussed in Section~\ref{sec:eval:autonomy_stack}).
To assess model performance, we express risk as the percentage of time the robot traversed a potentially dangerous terrain, 
$\lambda$, throughout its trajectory,
\begin{equation} \label{metric_risk_eval}
    Risk(\xi) =\sum_{s_i \in \xi}\frac{\lambda_{s_i}}{{|\xi|}}
\end{equation}
where ${|\xi|}$ represents the length of the trajectory and $\lambda_{s_i}$ is an indicator function returning $1$ when $\phi(s_i)$ contains the dangerous feature $\lambda$ and $0$ otherwise.
Assuming the unseen terrain's true reward is low or even negative, lower risk values should be correlated with the robot's safety.

Since risk alone is not a sufficient metric, we assess the overall performance as the tradeoff between risk and path length because longer paths tend to correspond with increased energy use and time to complete a mission.
\section{Evaluation} \label{sec:experiment}
This section evaluates RABRL in a 3-D simulated environment for an autonomous navigation task using a ground robot equipped with virtual sensors.

\subsection{Experiment Overview} \label{sec:eval:overview}

\begin{figure}[htbp]
\centerline{
    \includegraphics[scale=0.62]{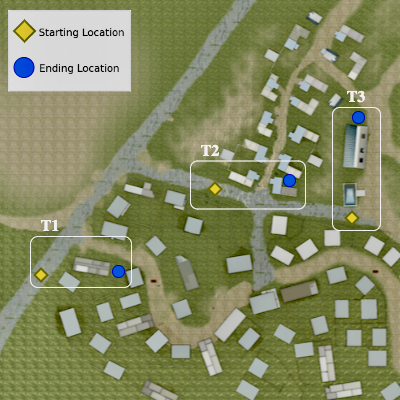}
 }
\caption{Birds-eye view of test scenarios in the simulated environment. Terrain colors are as follows: green corresponds to grass, brown to mud, and grey to asphalt.}
\label{fig_phx_env}
\end{figure}

All experiments were performed in a simulated 3-D Unity environment that represents a semi-structured village containing three terrain types describing our features, $\phi$, namely \emph{grass}, \emph{asphalt} and \emph{mud}, as well as several obstacles, including buildings, trees, and vehicles. Fig.~\ref{fig_phx_env} provides a birds-eye view of the three terrain types and buildings in the environment.

\noindent This simulated environment possesses greater complexity than the toy illustration shown in Fig.~\ref{fig:gridworld}, including: (i) noisy demonstrations due to imperfect perception and mapping, and (ii) the requirement to plan kinematically feasible trajectories.

To facilitate objective comparison, each reward model learning approach was trained with the same set of demonstrations, which were collected by having a human teleoperate the robot in an area of the simulated environment. When collecting these trajectories, the demonstrator stayed almost entirely on asphalt, in an attempt to show the robot their preference for driving on roads instead of grass. To showcase how the approaches handle learning from training data that lacks a full representation of the operating environment, demonstrations were collected in an area where no mud was present, thereby setting $\lambda$ to correspond to the mud feature. This methodology encompasses a number of real-world situations, including applications where training data cannot be collected in the precise operational environment or where there is significant time lapse or adverse weather conditions that cause an environment to change relative to the time at which training data was collected.

Three different reward models were trained, including (i) RABRL with a uninformed uniform prior, (ii) RABRL with an informed Dirichlet prior, and (iii) Maximum Entropy IRL~\cite{wigness_robot_2018}, which serves as a baseline towards learning semantic terrain rewards from human demonstrations, since both approaches use Maximum Entropy as a robot model.
Although the risk aversion idea from Hadfield-Menell et al.~\cite{hadfield-menell_inverse_2017} motivated this work, the methodology described there seeks to resolve a given misspecified, partially defined reward function.
As mentioned previously, RABRL seeks to learn a reward function solely from a dataset of demonstrations without ever being given a partially defined reward function, and is therefore difficult to compare directly.

We compare performance of each method on the three test scenarios
shown in Fig.~\ref{fig_phx_env}. To enable statistical analysis, including hypothesis testing, the robot started at the same location for each combination of test scenario and reward model, navigating to the same goal waypoint. Moreover, each combination of test scenario and reward model was run five times to account for stochasticity from imperfect mapping. The percentage of the time the robot went into the mud was calculated for each trajectory with Eq.~(\ref{metric_risk_eval}). Evaluation metrics are averaged across the five trial results.

\subsection{Robotic Autonomy Stack} \label{sec:eval:autonomy_stack}
\begin{figure}[!t]
\centerline{
    \includegraphics[scale=0.235]{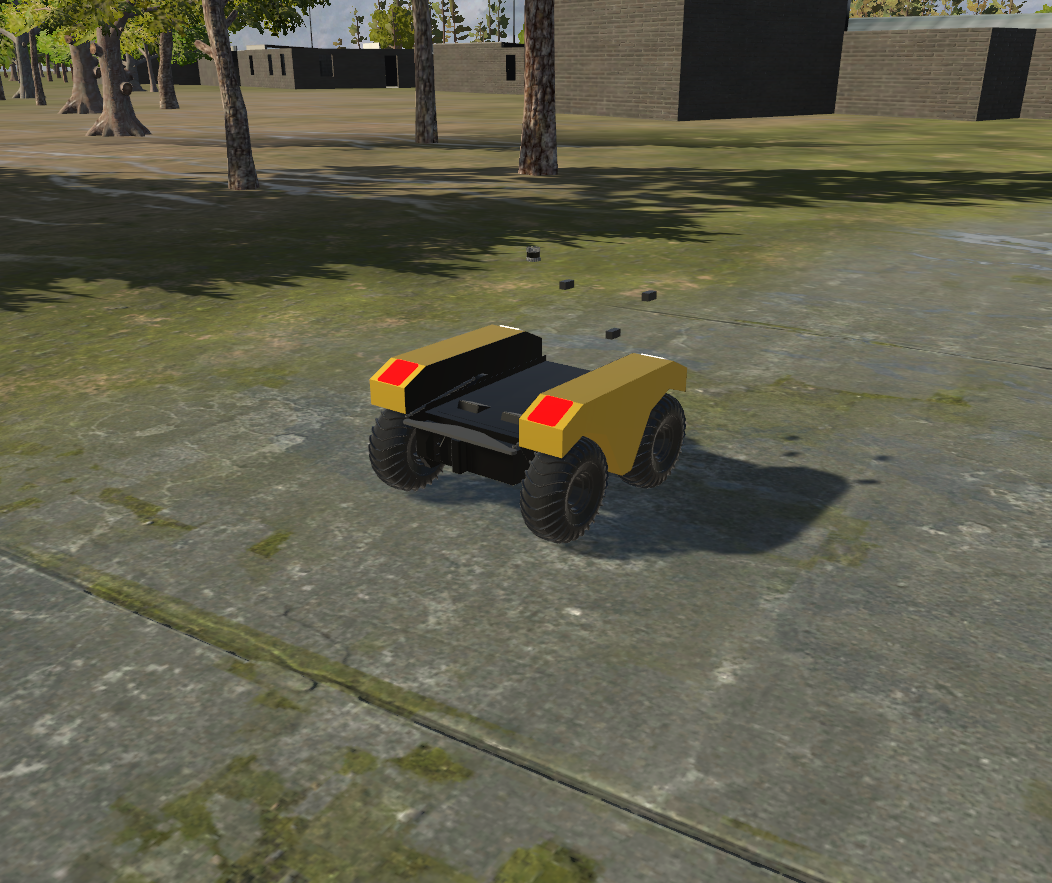}
}
\caption{Simulated environment with Clearpath warthog running ROS equipped with an array of virtual sensors.}
\label{fig_sim_warthog}
\end{figure}

A Clearpath Warthog equipped with an array of sensors including a 3D LiDAR, IMU, and two monocular cameras (Fig.~\ref{fig_sim_warthog}) was deployed for the  simulations. The robot possesses a full autonomy stack consisting of three main subsystems, namely mapping, perception, and planning. We briefly describe these subsystems, discussing how our contribution to risk-averse costmap generation interfaces with each of these subsystems.

The mapping subsystem is based on OmniMapper~\cite{Trevor14ICRA} and provides the necessary localization for autonomous navigation. The map from this subsystem represents a costmap layer for obstacle avoidance used by the planning subsystem. The risk-averse costmap acts as a second layer that provides terrain-awareness to the planning system, enhancing the overall navigation of the robot.
 
The perception subsystem includes semantic segmentation\footnote{For the experiments reported, we use the ground truth semantic segmentation produced by the simulation.} of camera images for an ontology of terrain and object classes such as grass, asphalt, and building. As images are segmented, the semantic label of each pixel is used to accumulate evidence for binary occupancy terrain grids, which represent the semantic features, $\phi$, used for reward model learning and risk-averse costmap generation.

As previously mentioned, the planning subsystem uses costmap layers generated from mapping and our IRL algorithm to plan paths during navigation. Specifically, costmaps serve as input to a global planner to search for the lowest-cost trajectory between the robot's location and a specified goal waypoint. In our system, global planning is computed with the Search-Based Planning Library \cite{sbpl} to find a kinematically achievable plan by searching combinations of motion primitives.

\subsection{Results} \label{sec:eval:results}
The dataset of demonstrations was used to train three different reward models.
If the robot were to operate fully online, imperfections from all other subsystems (perception, SLAM, mapping, and planning) would propagate.
Therefore, to best capture the performance of the proposed methodology, a robot collects a map of its environment a priori and then builds a costmap with respect to the learned reward weights. However, the reward function obtained from RABRL could be used to produce costmaps in an online setting.

For both of the RABRL reward models, Model (i) and (ii), a multivariate posterior was obtained from Eq.~(\ref{eq_posterior}).
The confidence parameter was set to $\beta = 0.3$ to indicate relatively low confidence in the optimality of the demonstrator and $\epsilon = 0.01$ was used, resulting in a threshold of $1 - 0.01 = 0.99$, indicating a relatively high tolerance to uncertainty.

The reward weight associated with a terrain was allowed to take on a value from the set
$\mathcal{W} = \{-2,...,1\}$. yielding a reward space of 64 possible reward functions, since $|\mathcal{W}|^\mathbb{D} =4^3 = 64$. This domain allowed each feature to be assigned a distinct value and also enabled a modest amount of reward scaling, since $|\mathcal{W}|$ was one larger than $\mathbb{D}$.
For the maximum entropy reward model, training was performed according to the reward model outlined in~\cite{wigness_robot_2018}.

\begin{table}[htbp]
    \caption{Marginal Entropy}
\begin{center}
    \begin{tabular}{|l|l|l|l|}
    \hline
    \multicolumn{1}{|c|}{\textbf{Reward Model}} & \multicolumn{3}{c|}{\textbf{Entropy - $H(w_i)$}} \\ \hline
    \rule{0pt}{2.2ex} \cellcolor[HTML]{9B9B9B} & \multicolumn{1}{c|}{$H(w_{grass})$} & \multicolumn{1}{c|}{$H(w_{mud})$} & \multicolumn{1}{c|}{$H(w_{asphalt})$} \\ \hline
    (i) RABRL w/ Unifrom                        & $0.974$                    & $0.981$                            & \rule{0pt}{2.2ex} $4.366 * 10^{-14}$   \\ \hline
    (ii) RABRL w/ Dirichlet                     & $0.759$                    & $0.797$                            & \rule{0pt}{2.2ex} $2.781 * 10^{-15}$   \\ \hline
    \end{tabular}
\label{table_phx_entropy}
\end{center}
\end{table}

Table \ref{table_phx_entropy} shows the normalized Shannon entropy, as evaluated by Eq.~(\ref{shannon_entropy}), of each marginal distribution obtained from Eq.~(\ref{eq:marginal}) for each terrain. Table \ref{table_phx_entropy} indicates that the normalized marginal entropy for mud was highest in both models, while the grass was second highest because the human demonstrations attempted to avoid grass.
However, both models are virtually certain of the preference for asphalt, as indicated by an entropy level close to zero.
Moreover, the entropy of Model (i) was higher because it employed an uninformed prior, whereas an informed prior, such as the Dirichlet prior in Model (ii), exhibited less uncertainty over the reward weights for each terrain. If the risk acceptance parameter $\epsilon$ was larger, resulting in less tolerance for risk, the respective reward weights would change. Specifically $\epsilon = 0.05$ results in weights of $-2$ for both grass and mud in the uniform model, thereby exemplifying the importance of an informed prior.

\begin{table}[htbp]
\caption{Feature Weight Vectors}
\begin{center}
    \begin{tabular}{|l|r|r|r|}
    \hline
    \multicolumn{1}{|c|}{\textbf{Reward Model}}     & \multicolumn{3}{c|}{\textbf{Feature Weight}} \\ \hline
    \cellcolor[HTML]{9B9B9B}{\color[HTML]{333333} } & \textbf{$w_{grass}$}   & \textbf{$w_{mud}$}  & \textbf{$w_{asphalt}$}  \\ \hline
    (i) RABRL w/Uniform                             & -0.258        & -0.687       & 1.000        \\ \hline
    (ii) RABRL w/Dirichlet                          & -0.687        & -1.253      & 1.000        \\ \hline
    (iii) Maximum Entropy IRL                       & -0.304        & 0.000        & 0.567        \\ \hline
    \end{tabular}
\label{table_phx_vector_weights}
\end{center}
\end{table}

Table ~\ref{table_phx_vector_weights} shows the estimated reward weight vector $\hat{w}=\langle\emph{grass}, \emph{mud}, \emph{asphalt}\rangle$ corresponding to the terrain features, which was determined with the weight selection technique described in Section \ref{sec:meth:weight_selection}. Model (iii) implicitly assigns the mud a reward weight of zero, thereby indicating a preference for mud over grass. Since mud was never encountered during training, its gradient during maximum likelihood optimization is always zero. Conversely, Models (i) and (ii) prefer every other terrain more than mud due to its high uncertainty.

\begin{table}[htbp]
\caption{Total Average Risk Taken}
\begin{center}
    \begin{tabular}{|l|c|r|r|r|}
    \hline
    \multicolumn{1}{|c|}{\textbf{Reward Model}} & \multicolumn{3}{c|}{\textbf{Total Risk}}                   \\ \hline
    \cellcolor[HTML]{9B9B9B}                    & \bf{T1}           & \bf{T2}               & \bf{T3}        \\ \hline
    (i) RABRL w/Uniform                        & 0.2090            & \bf{0.0000}          & \bf{0.0000}     \\ \hline
    (ii) RABRL w/Dirichlet                     & \bf{0.0617}       & \bf{0.0000}          & \bf{0.0000}     \\ \hline
    (iii) Maximum Entropy IRL                   & 0.3120            & 0.1724               & 0.8992          \\ \hline
    \end{tabular}
\label{table_phx_risk}
\end{center}
\end{table}

Table~\ref{table_phx_risk} shows the average risk, which was computed by averaging the values computed with Eq.~(\ref{metric_risk_eval}) from the five runs for each combination of test scenario and reward model. Table~\ref{table_phx_risk} indicates that the proposed method, RABRL, achieved lower risk in each test scenario and, in scenarios T2 and T3, the robot found a path to traverse which avoided mud entirely. Moreover, Table~\ref{table_phx_risk} shows that, in T1,  the informed Dirichlet prior (Model (ii)) accumulated approximately one third of the risk of Uniform prior (Model (i)) and one approximately one fifth of the risk of Maximum Entropy IRL (Model (iii)). Furthermore, while the informed prior took less risk than  its uninformative counterpart, scenarios T2 and T3 show that it is also possible to avoid side effects with uninformative priors.

\begin{figure}[htbp]
\centerline{
    \includegraphics[scale=0.395]{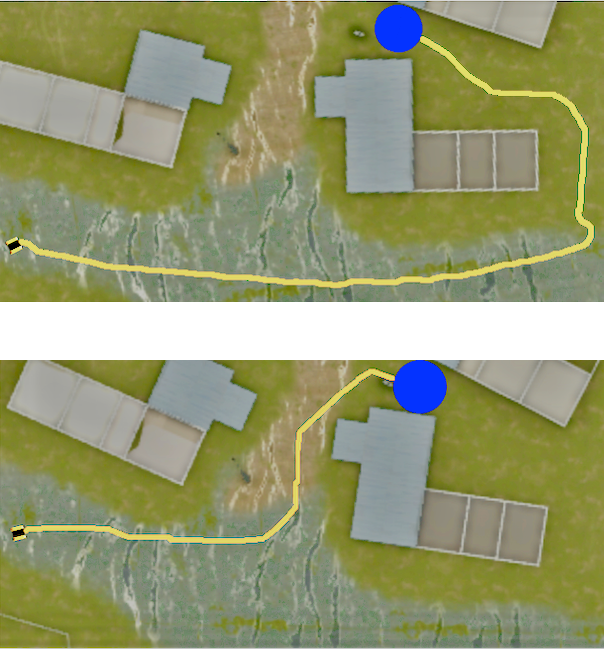}
}
\caption{Trajectory for Test Scenario 2 (T2) generated using two reward models with waypoint region (blue circle). \\ \textbf{Top:} (i) RABRL w/ Uniform  \ \ \ \ \textbf{Bottom:} (iii) MaxEnt IRL}
\label{fig_env_test_scenario}
\end{figure}

To further clarify the observations made in  Table~\ref{table_phx_risk}, Fig.~\ref{fig_env_test_scenario} shows a birds eye view of the trajectories taken by Models (i) and (iii) in T2. Model (iii) took a shorter path traversing mud, while Model (i) took a longer path and never traversed mud. Therefore, in certain scenarios RABRL was able to avoid negative side effects, which occur due to distributional shift in environmental terrain.

Table~\ref{table_phx_avg_path_len} shows the average path length computed with the five runs on each combination of test scenario and reward model. Table~\ref{table_phx_avg_path_len} indicates that Model (iii) took the shortest path in Test Scenarios T1 and T2. However, the path lengths of Models (i) and (ii) were competitive with Model (iii) in T1 and T2 took substantially more risk as noted in Fig.~\ref{fig_env_test_scenario}. Thus, in some scenarios, RABRL is able to substantially lower or eliminate risk while preserving a low path length. Moreover, in T3 Model (i) outperformed Model (iii) with respect to both total average risk and path length.

\begin{table}[htbp]
\caption{Average Path Length}
\begin{center}
    \begin{tabular}{|l|r|r|r|}
    \hline
    \multicolumn{1}{|c|}{\textbf{Reward Model}}     & \multicolumn{3}{c|}{\textbf{Test Scenario}} \\ \hline
    \cellcolor[HTML]{9B9B9B}{\color[HTML]{333333} } & \bf{T1}            & \bf{T2}           & \bf{T3}           \\ \hline
    (i) RABRL w/Uniform                                & 335.8         & 436.0          & \bf{454.6}        \\ \hline
    (ii) RABRL w/Dirichlet                              & 343.4         & 431.0          & 492.6        \\ \hline
    (iii) Maximum Entropy IRL                             & \bf{327.0}           & \bf{267.2}        & 465.0          \\ \hline
    \end{tabular}
\label{table_phx_avg_path_len}
\end{center}
\end{table}

\begin{figure}[!h]
\centerline{
    \includegraphics[scale=0.49]{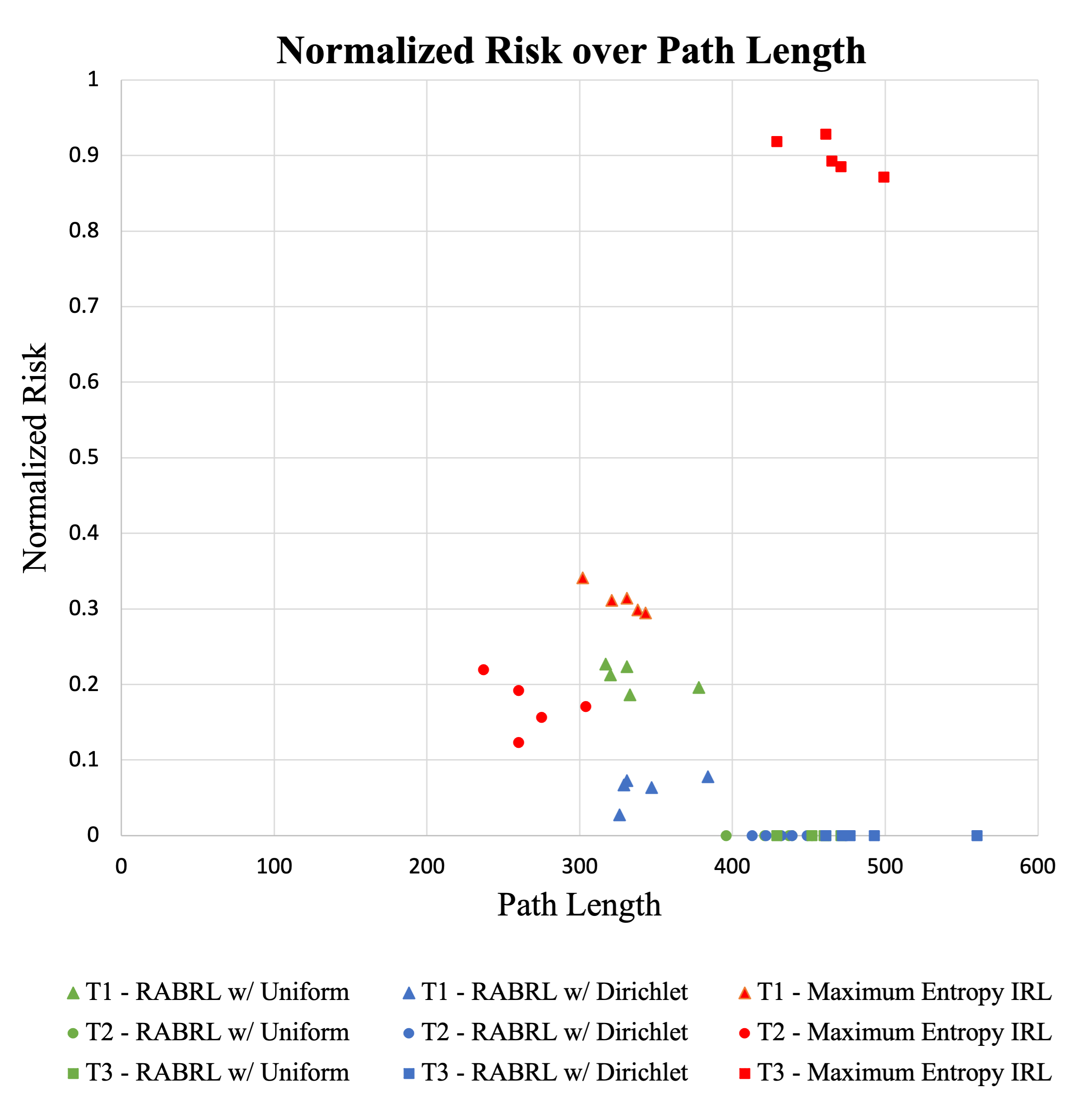}
}
\caption{Tradeoff between risk and path length.}
\label{fig_scatter}
\end{figure}

Fig.~\ref{fig_scatter} shows a scatter plot of the tradeoff between normalized risk (Eq.~(\ref{metric_risk_eval})) and path length, include all five runs for each combination of test scenario and reward model. To rigorously illustrate that RABRL reduced risk substantially, we performed a two means test on the risk taken for pairs of models. For example, a two-tailed test with a null hypothesis of equal risk in Models (ii) and (iii) was rejected in all three scenarios at the $99.95\%$ confidence level with p-values of $3.046034\times 10^{-8}$, $0.000449$, and $1.122941\times 10^{-7}$ respectively, strongly favoring RABRL. Furthermore, we performed a two means test on the path lengths for pairs of models In some cases, Model (iii) performed best, but in others the results were equivocal. Specifically, a two-tailed test with a null hypothesis of equal path lengths in Models (ii) and (iii) produced p-values of $0.247475$, $1.334867\times 10^{-5}$, and $0.227629$ for the three scenarios, suggesting that the null hypothesis of equal path length could not be rejected at the $90\%$ or even $80\%$ confidence level in Scenarios T1 and T3. Thus, while RABRL took a longer path to avoid mud in T2 as was shown in  Fig.~\ref{fig_env_test_scenario}, the lower risk taken by RABRL demonstrated very strong statistical significance without a significant increase in path length in two of three scenarios.

\section{Conclusion} \label{sec:conclusion}

This paper proposed a Bayesian technique to express the uncertainty over the semantic terrain reward weights of a linear reward function obtained from a dataset of human demonstrations. With the use of normalized Shannon Entropy, the relative uncertainty over reward weights can be learned by considering a small space of reward functions.
Experiments performed in a simulated 3D environment showed that a robot may leverage its uncertainty over semantic terrains to choose a trajectory with less risk.
However, this may require longer trajectories in some scenarios. Our proposed methodology, RABRL, enables an agent to avoid potentially dangerous terrain, while operating in an altered training environment.

The proposed technique is a member of the growing class of imitation learning techniques, which explicitly seek to avoid negative side effects that occur as a result of distributional shift of operational environments. Safe semi- or unstructured ground autonomy is likely to contain terrain scenarios never encountered during  training. Rather than undertake the infeasible task of attempting to capture such training data, a proactive approach to quantify uncertainty will identify gaps in training data and adapt behavior appropriately. By resolving ambiguities, implicit biases, and misspecifications in reward models obtained from human demonstrations, robots will be able to make more informed decisions, leading to safer behavior relative to their predecessors.

Future work includes exploring the potential application of RABRL to non-terrain related features of an environment that impose risk. Furthermore the use of MCMC methods may be able to scale the reward space $\mathcal{R}$, enabling larger scaled differences between features. Similarly, our method requires a feature indicator for a terrain whose reward weight is unknown. Future work could include using non-linear models on raw sensor data for feature extraction.

\bibliographystyle{./bibliography/IEEEtran}
\bibliography{root}

\clearpage
\end{document}